\documentclass[lettersize,journal]{IEEEtran}
\usepackage{times}

\usepackage[numbers]{natbib}
\usepackage{multicol}
\usepackage[bookmarks=true, colorlinks, linkcolor=black, menucolor=black, citecolor=black, urlcolor=blue]{hyperref}
\usepackage{caption}
\usepackage{subcaption}
\usepackage{xcolor}
\usepackage{multirow}
\usepackage{booktabs}
\usepackage{graphicx}
\usepackage{float} 
\usepackage{pifont}
\usepackage{natbib}
\usepackage{amsmath}
\usepackage{mathtools}
\usepackage{amssymb}
\usepackage{unicode-math}
\usepackage{makecell}
\usepackage{cuted}
\usepackage{capt-of}
\usepackage{url}

\begin{document}

\title{\LARGE \bf
Co-jump: Cooperative Jumping with Quadrupedal Robots via Multi-Agent Reinforcement Learning}


\author{Shihao Dong$^{*1}$, Yeke Chen$^{*1}$, Zeren Luo$^{1}$, Jiahui Zhang$^{1}$, Bowen Xu$^{1}$,\\
Jinghan Lin$^{1}$, Yimin Han$^{1}$, Ji Ma$^{1}$, Zhiyou Yu$^{2}$, Yudong Zhao$^{\dagger2}$, Peng Lu$^{\dagger1}$
\thanks{}
\thanks{$^{*}$ Equal Contribution. $^{\dagger}$Corresponding author: \url{lupeng@hku.hk} }
\thanks{$^{1}$ Adaptive Robotic Controls Lab (ArcLab), Department of Mechanical Engineering, University of Hong Kong, Hong Kong SAR, China, \url{dongsh24@connect.hku.hk}. }
\thanks{$^{2}$ EngineAI, Shenzhen, China, \url{zhaoyd@engineai.com.cn}. }
}

\maketitle

\begin{strip}
    \centering
    \vspace{-8.0em} 
    
    \includegraphics[width=1.0\textwidth]{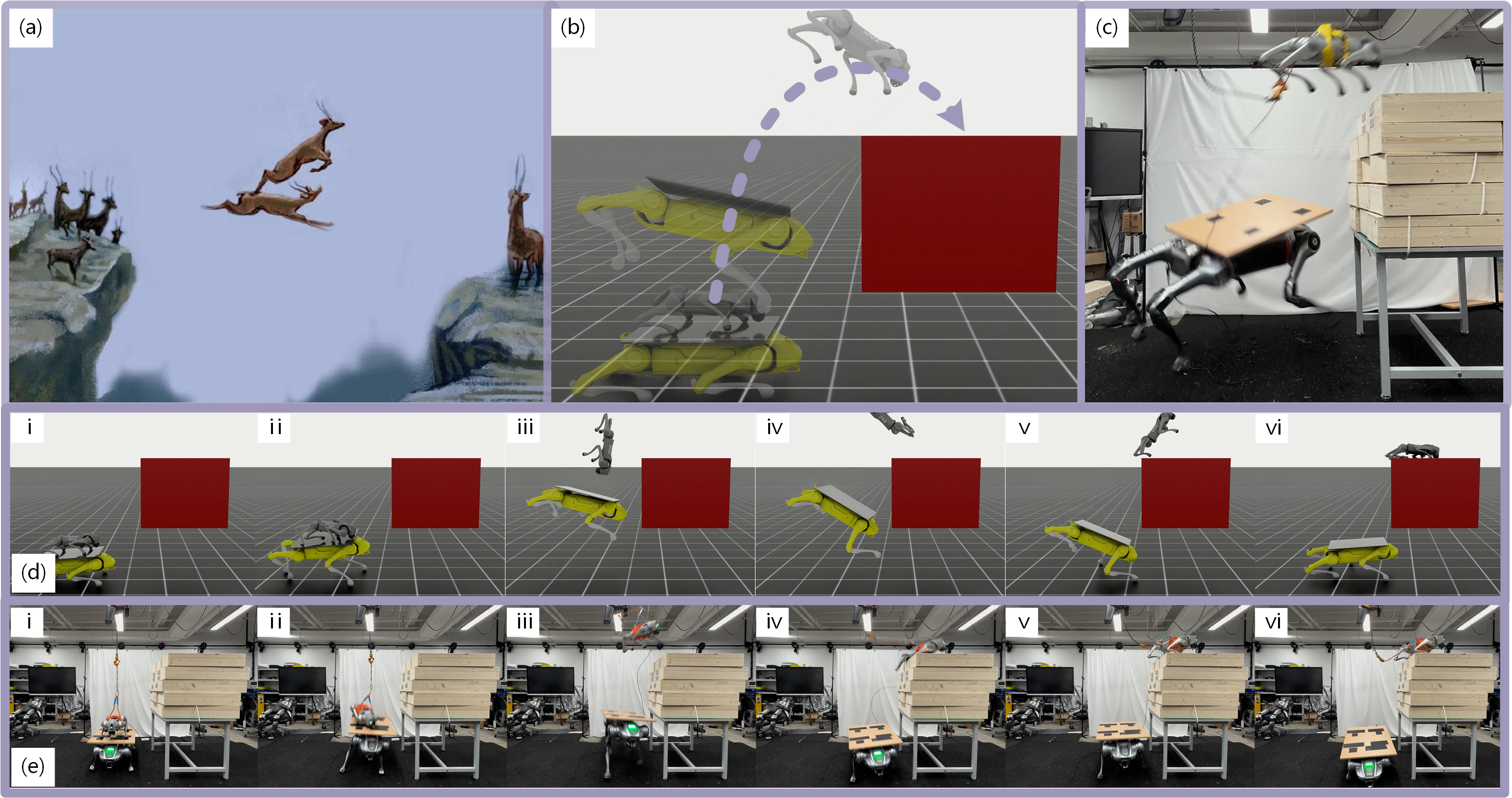}
    \captionof{figure}{The proposed framework enables cooperative leaping, allowing robots to traverse terrains far beyond the capability of a single robot. (a) Cooperative leaping behavior inspired by the "Flying Goral"~\cite{shen2016flying}; (b) forward Co-jump in simulation; (c) physical realization of Co-jump using two quadrupedal robots to reach a 1.5\,m platform; (d) sequence diagram of a front aerial flip; (e) sequence diagram of a side Co-jump achieving a height of 1.5\,m.}
    \label{pics:cover}
    \vspace{-1.0em} 
\end{strip}
\begin{abstract}

While single-agent legged locomotion has witnessed remarkable progress, individual robots remain fundamentally constrained by physical actuation limits. To transcend these boundaries, we introduce \textbf{Co-jump}, a cooperative task where two quadrupedal robots synchronize to execute jumps far beyond their solo capabilities. We tackle the high-impulse contact dynamics of this task under a decentralized setting, achieving synchronization without explicit communication or pre-specified motion primitives. Our framework leverages Multi-Agent Proximal Policy Optimization (MAPPO) enhanced by a progressive curriculum strategy, which effectively overcomes the sparse-reward exploration challenges inherent in mechanically coupled systems. We demonstrate robust performance in simulation and successful transfer to physical hardware, executing multi-directional jumps onto platforms up to \textbf{1.5 m} in height. Specifically, one of the robots achieves a foot-end elevation of 1.1 m, which represents a 144\% improvement over the 0.45 m jump height of a standalone quadrupedal robot, demonstrating superior vertical performance. Notably, this precise coordination is achieved solely through proprioceptive feedback, establishing a foundation for communication-free collaborative locomotion in constrained environments.

\end{abstract}

\IEEEpeerreviewmaketitle

\vspace{-1.0em}
\section{Introduction}

Despite remarkable progress in quadrupedal locomotion — enabling diverse locomotion gaits~\cite{han2024lifelike, yang2020multiexpert}, dynamic bipedal balancing~\cite{xiao2025learning, xiao2025stable}, and robust traversal of unstructured terrain~\cite{kumar2021rmaa, gangapurwala2022rloc, miki2022learning, zhuang2023robot, hoeller2024anymal, choi2023learning, kim2025highspeed, luo2024pie, dong2025marg} — the capabilities of individual legged robots remain fundamentally bounded by their physical scale, actuation limits, and energetic constraints. Consequently, tasks that exceed these intrinsic limits are physically unrealizable by a single agent, regardless of control sophistication. This \textbf{insurmountable physical barrier} raises a fundamental question: Can the locomotion limitations of a solitary robot be overcome through the collective potential of a multi-robot system?

Multi-robot collaboration has been explored in various contexts, such as dexterous manipulation~\cite{chen2022humanlevela, chen2024bidexhandsa} and hybrid locomotion-manipulation tasks with quadrupeds~\cite{an2025collaborative, feng2025learning, pandit2025multiquadruped}. However, cooperative dynamic locomotion remains largely unexplored. Inspired by the cooperative leaping behavior of gorals (Fig.~\ref{pics:cover}a), we introduce \textbf{Co-jump}: a novel asymmetric task where one quadruped transcends its solo capabilities by leveraging dynamic support from another. Specifically, a larger “launcher” (\textbf{Robot L}) actively exerts ground reaction forces to propel a smaller “jumper” (\textbf{Robot J}), effectively functioning as an active springboard. Robot J exploits this impulse to reach elevated platforms inaccessible to either robot individually. This maneuver demands precise spatiotemporal synchronization and high-dynamic motion control, relying exclusively on proprioceptive feedback and precise motor actuation, without the aid of external perception or explicit trajectory planning.

Co-jump inherits the fundamental challenges of high-dynamic locomotion with a single-agent. Beyond these, it introduces additional complexities unique to multi-agent coordination. The joint action space grows significantly larger and more complex with the number of agents~\cite{huh2024multiagent}. This scalability challenge is further compounded by the credit assignment problem~\cite{ning2024survey}: as the team size grows, accurately evaluating each agent’s individual contribution from sparse, global task rewards becomes increasingly difficult, especially in asymmetric multi-agent configurations. Moreover, without explicit communication and relying solely on local observations, agents must achieve precise spatiotemporal coordination through physical interaction alone. This direct mechanical coupling between robots induces strongly non-stationary environmental dynamics~\cite{gronauer2022multiagent}, rendering the system highly sensitive to minor behavioral deviations.

To address these challenges, we introduce a reinforcement learning (RL) framework based on Multi-Agent Proximal Policy Optimization (MAPPO)\cite{yu2022surprising}. Utilizing a Centralized Training with Decentralized Execution (CTDE) architecture\cite{amato2024introduction}, our approach incorporates a shared reward mechanism alongside a progressive curriculum. This design enables agents to autonomously develop effective cooperative strategies without relying on reference trajectories or motion capture data. We showcase robust Co-jump behaviors in simulation (Fig.\ref{pics:cover}b) and extend the framework to challenging cooperative aerial flips (Fig.\ref{pics:cover}d). Most importantly, the learned policies are successfully transferred to robot hardware(Fig.\ref{pics:cover}c, e), achieving real-world demonstrations of the Co-jump maneuver. The main contributions of this work are listed here:
\begin{itemize} 
  \item We propose a communication-free MARL framework that enables high-dynamic, tightly coupled cooperative jumping relying exclusively on proprioception. This demonstrates that precise physical coordination can be achieved without visual perception or external trajectory planning. 
  \item We introduce a progressive curriculum learning strategy that enables the autonomous emergence of complex coordination policies from scratch, completely eliminating the reliance on reference trajectories or motion capture data. 
  \item We design multiple reward functions that align each robot's motor commands with shared collaborative goals, effectively resolving the multi-agent credit assignment problem inherent in asymmetric physical interactions and guiding agents toward better cooperation.
  \item We validate our approach through real-world experiments, achieving a platform clearance of 1.5\,m and demonstrating a 144\% improvement in foot-end elevation (1.1\,m) over the 0.45\,m baseline of conventional single quadrupeds～\cite{han2025omninet}.
\end{itemize}

\section{Related Works}
\subsection{Quadruped Robot Jumping Control}
Classical approaches to jumping control for quadrupedal robots typically rely on model-based methods, which use accurate dynamic models in combination with trajectory optimization or model predictive control to generate dynamically feasible jump trajectories~\cite{gilroy2021autonomous, nguyen2019optimized, song2022optimal, yue2024onlinea, nguyen2022contacttiming}. These methods can achieve high-precision performance in known and structured environments. However, their effectiveness often degrades in the presence of model mismatch or external disturbances~\cite{he2024agilea}.

Recent work has increasingly turned to learning-based control, particularly reinforcement learning, which enables quadrupeds to acquire agile jumping behaviors through direct interaction with the environment~\cite{kim2025stagewise, han2025omninet, chane-sane2024soloparkoura}. Within this paradigm, some approaches employ imitation learning, distilling jumping skills from expert demonstrations or simulation-generated reference trajectories~\cite{luo2025learning, zhao2025learning, zhang2024learninga, smith2023learninga}. However, trajectory-dependent learning approaches face significant practical and representational challenges. High-quality reference trajectories for complex, high-dynamic maneuvers are often unavailable, particularly when the task involves novel multi-agent interactions characterized by strong coupling. To address this, we propose a fully reference-free framework that does not rely on expert demonstrations. Our approach uses curriculum learning and shared reward shaping to enable agents to discover effective Co-jump strategies.

\subsection{Multi-Agent Reinforcement Learning for Robot Control}

MARL has become a common framework for coordinating multiple robots, with existing methods broadly falling into two categories. One widely used approach adopts a hierarchical control structure, where a high-level planner produces abstract coordination signals and delegates their execution to low-level policies that generate motor commands~\cite{pandit2025multiquadruped, an2025collaborative, feng2025learning, su2025realworld}. This decomposition offers modularity, simplifies the integration of prior knowledge and safety constraints, and often improves real-world transfer due to reduced learning complexity at the actuation level. However, in highly dynamic cooperative tasks like Co-jump, such architectures can introduce critical bottlenecks. The separation between planning and execution creates an information gap that prevents fine-grained coordination of impulsive contact forces and precise timing, both of which are essential during launch. 

An alternative paradigm in MARL for robot control employs end-to-end policies that directly map raw observations to low-level actions~\cite{gu2023safea, zhao2025exploiting, gao2024coohoi, baltes2025cooperativea}. By jointly optimizing the entire control pipeline, these methods can discover compact and tightly coupled coordination strategies that are difficult to encode through explicit planning abstractions. Motivated by these advances, we adopt an end-to-end MARL approach for Co-jump to circumvent the accumulated errors and representational bottlenecks inherent in hierarchical architectures.

\section{Method}
\begin{figure*}[!htbp]
   \centering
   \includegraphics[width=0.9\textwidth]{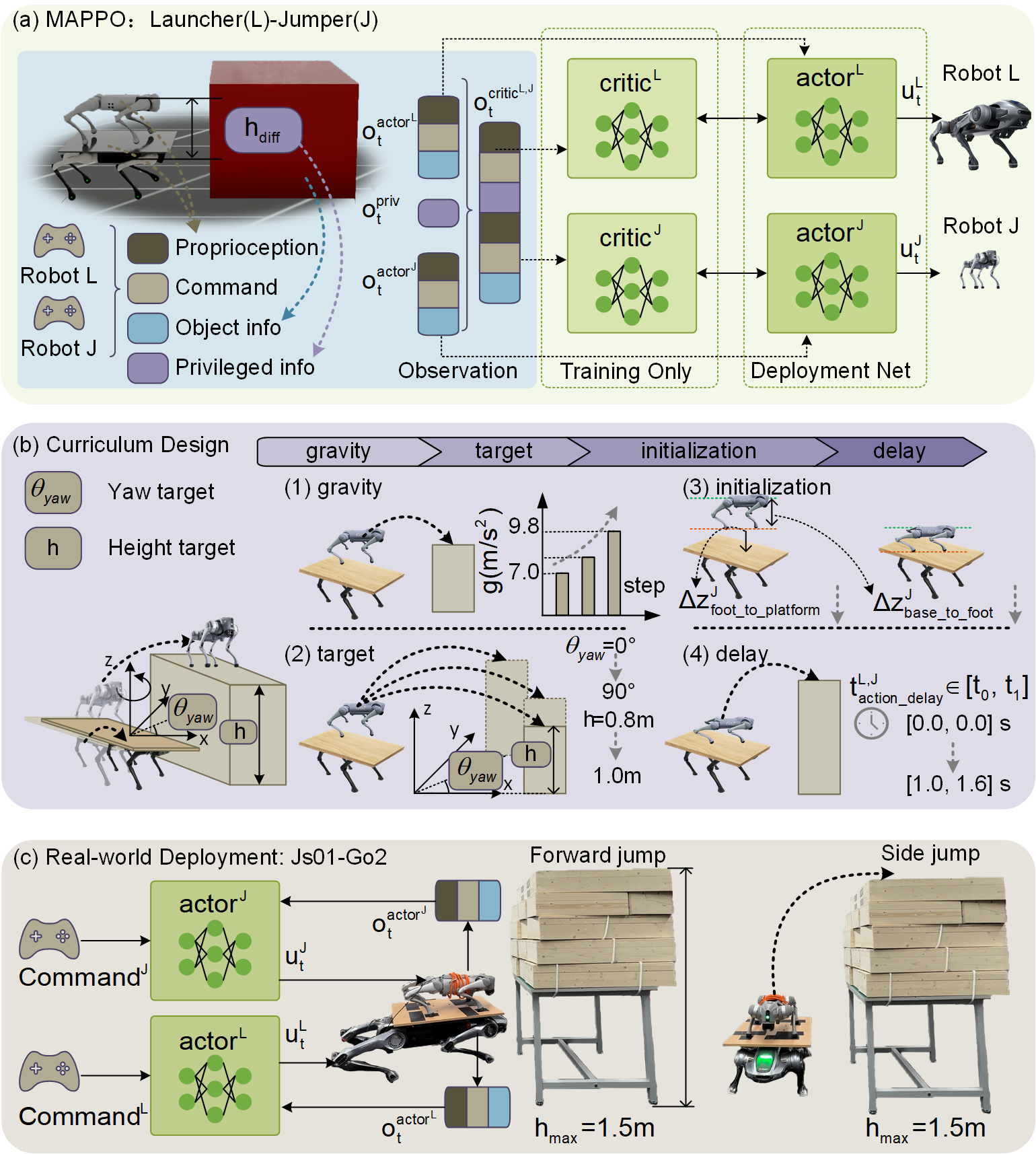}
   \caption{Overview of the Co-jump framework. (a)MAPPO architecture with independent policy and value networks for Robot L and Robot J. (b) Four-stage curriculum learning progression: gravity, target, initialization, and delay curricula. (c) Real-world implementation of Co-jump on quadrupedal robots.}
   \label{pics:framework}
   \vspace{-1.0em}
\end{figure*}

This section presents our MARL framework, which employs a four-stage curriculum learning approach as illustrated in Fig.~\ref{pics:framework}. We first introduce the problem formulation and core components in Section~\ref{problem}, followed by detailed descriptions of the reward design in Section~\ref{reward} and the curriculum learning strategy in Section~\ref{curriculum}. The overall framework integrates CTDE architecture, enabling effective multi-robot coordination for Co-jump tasks.

\subsection{Problem Statement} \label{problem}
The Co-jump problem is a cooperative task in which multiple quadrupedal robots must execute temporally synchronized jumps using only proprioceptive feedback, without access to exteroceptive sensing or inter-robot communication. Although coordination is required, each robot optimizes its own performance objective. We model this as a Decentralized Partially Observable Markov Decision Process (Dec-POMDP), defined by the tuple $\langle \mathcal{N}, \mathcal{S}, \mathcal{U}, \Omega, \mathcal{O}, \mathcal{P}, \mathcal{R}, \gamma \rangle$. Here, $\mathcal{N} = \{1, \dots, N\}$ indexes the set of $N$ agents, each sharing the same individual action space $\mathcal{U}$ and observation space $\Omega$. The global state $ \mathbf{s}_t \in \mathcal{S}$ fully describes the joint configuration of all robots as well as relevant environmental conditions. At each time step $t$, the agents execute a joint action $\mathbf{u}_t = (\mathbf{u}_t^1, \dots, \mathbf{u}_t^N) \in \mathcal{U}^N$, which induces a transition to the next global state $\mathbf{s}_{t+1}$ according to $\mathcal{P}(\mathbf{s}_{t+1} \mid \mathbf{s}_t, \mathbf{u}_t)$. Subsequently, a joint observation $\mathbf{o}_{t+1} = (\mathbf{o}_{t+1}^1, \dots, \mathbf{o}_{t+1}^N)$ is generated according to the observation function $\mathcal{O}(\mathbf{s}_{t+1}, \mathbf{u}_t)$, where each component $\mathbf{o}_{t+1}^i \in \Omega$ constitutes the local proprioceptive measurement available only to agent $i$.

The reward function $\mathcal{R}: \mathcal{S} \times \mathcal{U}^N \to \mathbb{R}^N$ assigns to each agent $i$ an individual scalar reward $\mathcal{R}^i(\mathbf{s}_t, \mathbf{u}_t) \in \mathbb{R}$ that reflects both its own jumping performance and implicit coordination with others. With discount factor $\gamma \in [0, 1)$, each agent learns a decentralized policy $\pi^i(\mathbf{u}_t^i \mid \mathbf{o}_t^i)$. The joint policy $\symbfup{\pi}=\{\pi^1, \dots, \pi^N\}$ induces trajectories $\tau = (\mathbf{s}_0, \mathbf{u}_0, \mathbf{s}_1, \mathbf{u}_1, \dots)$, and each agent $i$ aims to maximize its own expected cumulative discounted return:
\begin{equation}
  J^i(\symbfup{\pi}) = \mathbb{E}_{\tau \sim \symbfup{\pi}} \left[ \sum_{t=0}^{\infty} \gamma^t \mathcal{R}^i(\mathbf{s}_t, \mathbf{u}_t) \right].
\end{equation}

\subsubsection{Observation Space}
At each time step $t$, agent $i$ receives a proprioceptive observation vector $\mathbf{o}_t^i \in \mathbb{R}^{52}$, defined as
\begin{equation}
    \mathbf{o}_t^i = [\symbfup{\omega}_t^i, \mathbf{g}_t^i, \mathbf{q}_t^i, \dot{\mathbf{q}}_t^i, \mathbf{u}_{t-1}^i, \mathbf{s}_{\text{cmd}}^i, \mathbf{s}_{\text{obj}}],
\end{equation}
where $\symbfup{\omega}_t^i \in \mathbb{R}^3$ denotes the base angular velocity and $\mathbf{g}_t^i \in \mathbb{R}^3$ represents the gravity vector, both expressed in the base frame. The joint state is captured by the joint angles $\mathbf{q}_t^i \in \mathbb{R}^{12}$ and velocities $\dot{\mathbf{q}}_t^i \in \mathbb{R}^{12}$, along with the action $\mathbf{u}_{t-1}^i \in \mathbb{R}^{12}$ executed at the previous time step. The command $\mathbf{s}_{\text{cmd}}^i \in \mathbb{R}^4$ specifies the desired launch kinematics, consisting of the target horizontal velocity $\mathbf{v}_{\text{xy}}^i \in \mathbb{R}^2$, desired yaw target $\theta_{\text{yaw}}^i \in \mathbb{R}$, and reference base height $h_{\text{base}}^i \in \mathbb{R}$. Additionally, each agent observes $\mathbf{s}_{\text{obj}} \in \mathbb{R}^6$, which encodes the three-dimensional position and size of the shared target object.

For the centralized critic used during training, the global state is constructed by concatenating the local observations of both agents and their relative base position difference $\mathbf{p}^{\text{diff}}_t = \mathbf{p}_t^L - \mathbf{p}_t^J \in \mathbb{R}^3$, yielding $\mathbf{s}_t = [\mathbf{o}_t^L, \mathbf{o}_t^J, \mathbf{p}^{\text{diff}}_t]$. This state representation provides the critic with sufficient multi-agent and spatial information to resolve credit assignment in cooperative tasks, while the actor policies remain decentralized and rely only on local observations during execution.

\subsubsection{Action Space}
The action space for each agent is defined as $\mathbf{u}_t^i \in \mathbb{R}^{12}$, which parameterizes the deviation from a default joint configuration. The desired joint target $\mathbf{q}_t^{i*}$ is computed as $\mathbf{q}_t^{i*} = \widetilde{\mathbf{q}}_t^i + \beta \mathbf{u}_t^i$, where $\widetilde{\mathbf{q}}_t^i \in \mathbb{R}^{12}$ denotes the default joint pose and $\beta > 0$ is a scalar action scale that controls the magnitude of policy outputs. Low-level motor commands are generated using a standard proportional-derivative (PD) controller.

\subsection{Reward Function} \label{reward}
The reward function for each agent is composed of three complementary components: (1) \textbf{Task reward} that measures individual jumping performance, (2) \textbf{Regularization reward} that promotes physically plausible and smooth motion, and (3) \textbf{Cooperation reward}, shared across all agents, that incentivizes synchronized multi-robot behavior. The total reward at time step $t$ is given by the weighted sum:
\begin{equation}
r_t = \alpha_{\text{task}} \cdot r^{\text{task}}_{t} + \alpha_{\text{regu}} \cdot r^{\text{regu}}_{t} + \alpha_{\text{coop}} \cdot r^{\text{coop}}_{t},
\label{eq:reward_total}
\end{equation}
where $\alpha_{\text{task}}$, $\alpha_{\text{regu}}$, and $\alpha_{\text{coop}}$ are non-negative scaling coefficients.

Specifically, the task reward $r^{\text{task}}_t$ evaluates key aspects of individual jump execution, including tracking accuracy with respect to the desired motion command, correct foot contact state management, and other jump quality metrics. The regularization reward $r^{\text{regu}}_t$ penalizes behaviors that compromise physical feasibility or control smoothness—namely, large action derivatives, violations of joint position or velocity limits, deviations from an upright body orientation and abrupt kinematic changes. Finally, the cooperation reward $r^{\text{coop}}_t$, which is identical for all agents, encodes collective objectives: it maximizes inter-agent height differences during aerial phases, provides a shared bonus upon successful Co-jump completion, rewards proper termination conditions, and applies a common penalty if Robot J falls.

The specific terms and their corresponding weights are detailed in Table~\ref{reward_function}, with the component coefficients set to $\alpha_{\text{task}}=1$, $\alpha_{\text{regu}}=1$, and $\alpha_{\text{coop}}=1$.

To enforce state compliance within desirable bounds while preserving differentiable gradients, we utilize the tolerance function $f_{\text{tol}}$~\cite{huang2025learninga, tao2022learninga, tassa2018deepmind}. This function maps a scalar input $x$ to a utility value within $[0, 1]$, parameterized by a target interval $b=[b_l, b_u]$, a margin width $m > 0$, and a target value $v \in (0, 1)$ at the margin boundary. The general form is defined as:

\begin{equation}
f_{\text{tol}}(x, b, m, v) = \begin{cases} 
1 & \text{if } b_l \leq x \leq b_u, \\
g\left(\frac{b_l - x}{m}\right) & \text{if } x < b_l, \\
g\left(\frac{x - b_u}{m}\right) & \text{if } x > b_u,
\end{cases}
\end{equation}
where $g(\cdot)$ is a shaping kernel determining the decay profile outside the boundary. We employ a long-tail kernel~\cite{tassa2018deepmind}:
\begin{equation}
g(z) = \frac{1}{\left(z \cdot \sqrt{\frac{1}{v} - 1}\right)^2 + 1},
\end{equation}
where $z$ represents the normalized deviation from the bounds (i.e., $z = \frac{\Delta x}{m}$). This formulation ensures that when the deviation equals the margin $m$ ($z=1$), the reward value decays exactly to $v$.

\begin{figure}[!htbp]
   \centering
    \includegraphics[width=0.4\textwidth]{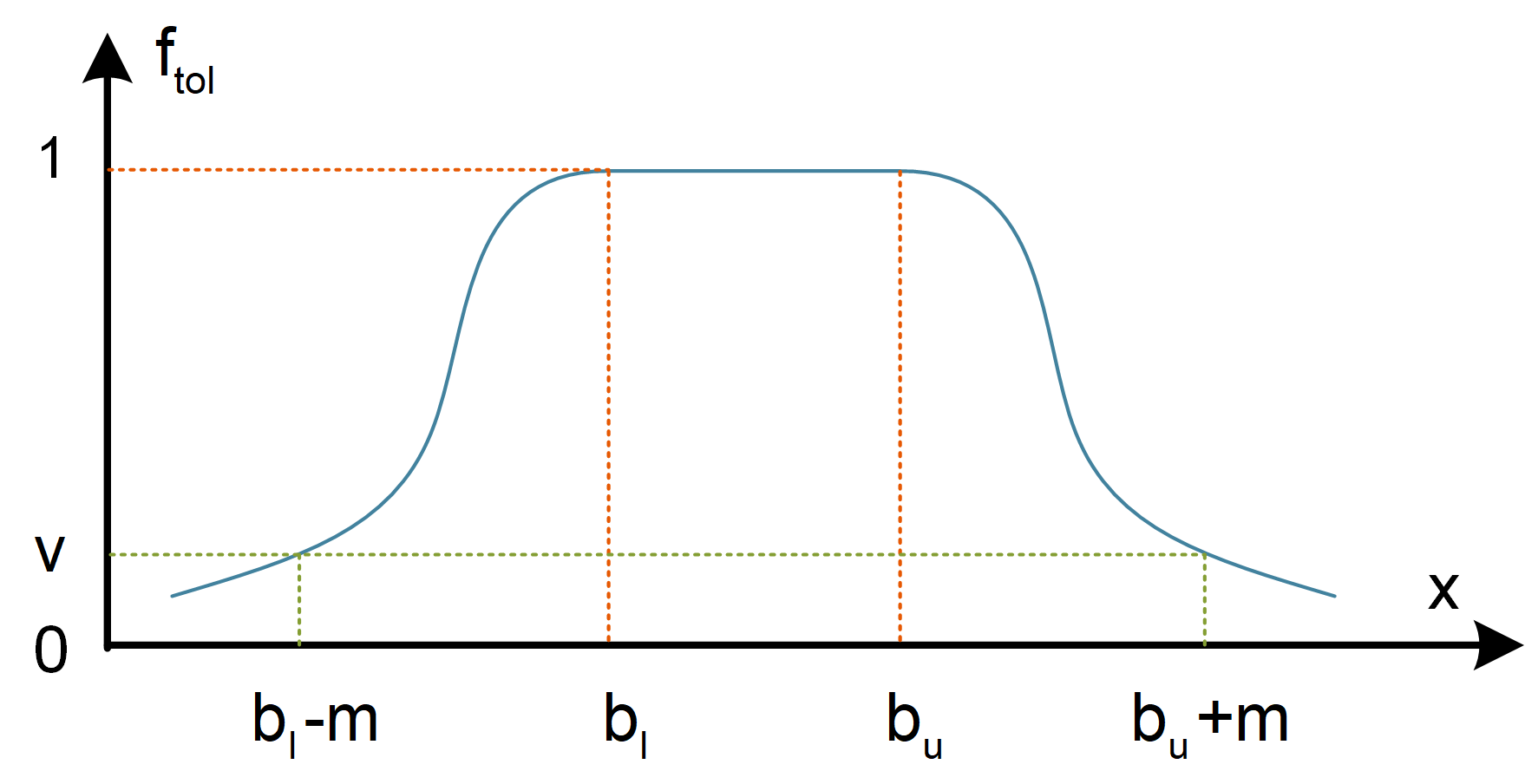}
   \caption{Illustration of the tolerance function $f_{\text{tol}}$. The reward is maximized within the bounds $[b_l, b_u]$ and decays smoothly based on the margin $m$ and value $v$~\cite{tao2022learninga}.}
   \label{pics:ftol}
   \vspace{-0.5em}
\end{figure}

\subsubsection{Task Phases and Success Criteria}

Following the curriculum-based framework~\cite{atanassov2025curriculumbased}, the jumping motion is decomposed into three sequential phases:
\begin{itemize}
    \item \textbf{Initial Phase:} All four legs are in contact with the ground, and the robot crouches in preparation for takeoff.
    \item \textbf{Flight Phase:} The aerial state where all feet are fully detached from the ground.
    \item \textbf{Landing Phase:} All limbs re-establish ground contact, and the robot transitions into a stable posture.
\end{itemize}

\paragraph{Standard Co-jump Success}
For the standard cooperative jumping task, a successful jump is defined by the touchdown status of Robot J:
\begin{equation}
  \label{eq:success}
  \mathbb{1}(\text{success}) = \mathbb{1}(\text{td})
\end{equation}
where $\mathbb{1}(\text{td})$ indicates a successful touchdown on the target platform, satisfying height, precision, and orientation constraints:
\begin{equation}
  \label{eq:touch_down}
  \begin{split}
  \mathbb{1}(\text{td}) = &\mathbb{1}(h^{J}_z > 1.0) \cdot \\
  \quad &\mathbb{1}(|p^{J}_{xy}-p^{\text{target}}_{xy}| < 0.4) \cdot \\
  \quad &\mathbb{1}(q^{J}_z < 0.0)
  \end{split}
\end{equation}
where $h^{J}_z$ is the base height, $p^{J}_{xy}$ is the platform position, and $q^{J}_z$ denotes the projected gravity vector ensuring an upright posture.

\paragraph{Flip Maneuver Success}
For the advanced flip curriculum, the success condition is augmented to require a full front flip. Robot J must accumulate a specific pitch rotation while maintaining directional stability:
\begin{equation}
  \label{eq:success_flip}
  \begin{split}
  \mathbb{1}(\text{success}_{\text{flip}}) = &\mathbb{1}(\text{td}) \cdot \mathbb{1}(\varphi_{\text{acu\_pitch}}^{J} > \frac{3\pi}{2}) \cdot \mathbb{1}(\text{ori})
  \end{split}
\end{equation}
Here, $\varphi_{\text{acu\_pitch}}^{J}$ represents the accumulated pitch angle, and $\mathbb{1}(\text{ori})$ enforces strict directional control to prevent deviation during the flip:
\begin{equation}
  \label{eq:orientation}
  \mathbb{1}(\text{ori}) = \mathbb{1}(|\varphi_{\text{roll}}^{J}|<\frac{\pi}{6}) \cdot \mathbb{1}(|\varphi_{\text{yaw}}^{J}|<\frac{\pi}{6})
\end{equation}

\begin{table*}[h]
\caption{Detailed reward terms for the training policy. $C_{\text{part}}$ denotes the contact force magnitude on a specific body part. $FL, FR, RL, RR$ denote Front-Left, Front-Right, Rear-Left, and Rear-Right, respectively.}
\vspace{-0.8em}
\label{reward_function}
\setlength\tabcolsep{3pt}  
\begin{center}

\resizebox{\textwidth}{!}{ 
\begin{tabular}{l c c l l}
\hline
\multirow{2}{*}{\textbf{Term}} & \multicolumn{2}{c}{\textbf{Weight}} & \multirow{2}{*}{\textbf{Equation}}  & \multirow{2}{*}{\textbf{Description}} \\
\cline{2-3}\\[-1.5ex]
 & \makecell{\textbf{Robot L}\\ \textbf{(Js01/Aliengo)}} & \makecell{\textbf{Robot J}\\ \textbf{(Go1/Go2)}} & & \\[0.2ex]
\hline
\\ [-1.5ex]
\multirow{14}*{$r^{\text{regu}}_{t}$} 
  & -6.0 & -6.0 & $\mathbb{1}(|\varphi_{\text{roll}}|>\frac{\pi}{12})$  & Roll angle deviation penalty\\ [+0.3ex]
  & -6.0 & -6.0 & $\mathbb{1}(|\varphi_{\text{yaw}}|>\frac{\pi}{12})$   & Yaw angle deviation penalty\\ [+0.3ex]
  & $-2.5\times10^{-7}$ & $-2.5\times10^{-7}$ & $\|\ddot{\mathbf{q}}\|^2$    & Joint acceleration penalty\\ [+0.3ex]
  & $-1.0\times10^{-4}$ & $-1.0\times10^{-4}$ & $\|\dot{\mathbf{q}}\|^2$    & Joint velocity penalty\\ [+0.3ex]
  & $-2.5\times10^{-6}$ & $-2.5\times10^{-5}$ & $\|\symbfup{\tau}\|^2$    & Joint torque penalty\\ [+0.3ex]
  & -0.5 & -0.5 & $\|\mathbf{q}-\mathbf{q}_{\text{target}}\|^2$    & Joint tracking penalty\\ [+0.3ex]
  & -0.1 & -0.1 & $\|\mathbf{u}_t-2\mathbf{u}_{t-1}+\mathbf{u}_{t-2}\|^2$    & Action smoothness penalty\\ [+0.3ex]
  & -0.1 & -0.1 & $\|\mathbf{u}_t-\mathbf{u}_{t-1}\|^2$    & Action rate penalty\\ [+0.3ex]
  & -0.5 & -0.5 & $\|\mathbf{q}_{\text{hip}}-\mathbf{q}_{\text{hip}}^{\text{default}}\|^2$    & Hip joint deviation penalty\\ [+0.3ex]
  & -0.2 & -0.2 & $\|\mathbf{q}_{\text{thigh}}-\mathbf{q}_{\text{thigh}}^{\text{default}}\|^2$    & Thigh joint deviation penalty\\ [+0.3ex]
  & -0.2 & -0.2 & $\|\mathbf{q}_{\text{calf}}-\mathbf{q}_{\text{calf}}^{\text{default}}\|^2$    & Calf joint deviation penalty\\ [+0.3ex]
  & -0.8 & -0.8 & $\mathbb{1}(C_{\text{calf}} > 1)$    & Calf contact penalty\\ [+0.3ex]
  & -0.8 & -0.8 & $\mathbb{1}(C_{\text{thigh}} > 1)$    & Thigh contact penalty\\ [+0.3ex]
  & -0.8 & -0.8 & $\mathbb{1}(C_{\text{base}} > 1)$    & Base contact penalty\\ [+0.3ex]
  & -5.0 & - & $\|\mathbf{q}_{xy}\|^2$    & Upright posture penalty \\ [+0.3ex]
  & - & -2.0 & $\|\mathbf{q}_{xy}\|^2 \cdot \mathbb{1}(\text{success})$    & Upright posture penalty after successful jumping\\ [+0.3ex]
\hline
\\ [-1.5ex]
\multirow{9}*{$r^{\text{task}}_{t}$} 
  & 8.0 & 8.0 & $f_{\text{tol}}(h_{\text{base}}-h_{\text{base}}^*, [0,\infty], 0.2, 0.2) \cdot \mathbb{1}(\text{flight})$ & Command height tracking (flight)\\ [+0.3ex]
  & 10.0 & 10.0 & $f_{\text{tol}}(\mathbf{v}_{xy}-v_{xy}^*, [-\infty,0.2], 0.5, 0.2) \cdot \mathbb{1}(\text{flight})$ & Command $\mathbf{v}_{xy}$ tracking (flight)\\ [+0.3ex]
  & 2.0 & 2.0 & $f_{\text{tol}}(\omega_{\text{yaw}}-\omega_{\text{yaw}}^*, [-\infty,0.1], 0.2, 0.2) \cdot \mathbb{1}(\text{flight})$ & Desired $\omega_{\text{yaw}}$ tracking (flight)\\ [+0.3ex]
  & -2.0 & -2.0 & $||\mathbf{p}_{z}^{\text{feet}} + 0.15||^2  \cdot \mathbb{1}(\text{flight})$    & Feet height retraction reward (flight)\\ [+0.3ex]
  & 3.0 & 3.0 & $\mathbb{1}(C_{\text{feet}}) \cdot (\mathbb{1}(\text{initial} \lor \text{landing}))$    & Foot contact maintenance\\ [+0.3ex]
  & 2.0 & 2.0 & $f_{\text{tol}}(h_{\text{base}}-h_{\text{initial}}, [-\infty,0], 0.1, 0.5) \cdot \mathbb{1}(\text{initial})$    & Squatting encouragement (initial)\\ [+0.3ex]
  & -0.1 & -0.1 & $\mathbb{1}(C_{\text{foot,FL}} \neq C_{\text{foot,FR}} \lor C_{\text{foot,RL}} \neq C_{\text{foot,RR}})$    & Asymmetric feet contact penalty\\ [+0.3ex]
  & 2.0 & 2.0 & $\mathbb{1}(\text{landing}) \cdot \mathbb{1}(\text{success})$    & Post-landing stabilization\\ [+0.3ex]
\hline
\\ [-1.5ex]
\multirow{6}*{${r}^{\text{coop}}_{t}$} 
     & \multicolumn{2}{c}{6.0} & $f_{\text{tol}}(h^{J}-h^{L}, [0.6, \infty], 0.3, 0.2)$  & Height difference reward\\ [+0.3ex]
     & \multicolumn{2}{c}{-2.0} & $\mathbb{1}(\theta_{\text{pitch}}^{J}<0 \lor \theta_{\text{pitch}}^{J}>\pi/4)$ & Pitch deviation of Robot J penalty\\ [+0.3ex]
     & \multicolumn{2}{c}{40.0} & $\mathbb{1}(\text{success})$ & Success Reward\\ [+0.3ex]
     & \multicolumn{2}{c}{-2.0} & $\mathbb{1}(\|\mathbf{f}_{\text{base}}\|>1500)$    & Termination penalty\\ [+0.3ex]
     & \multicolumn{2}{c}{-2.0} & $\mathbb{1}(h_{\text{base}}^{J} < 0.4)$    & Falling of Robot J penalty\\ [+0.3ex]
\hline
\end{tabular}
} 
\end{center}
\vspace{-1.5em}
\end{table*}

\section{Curriculum Learning Design} \label{curriculum}

The Co-jump task demands tightly coordinated, high-impulse maneuvers between two legged robots under strong mechanical coupling, posing significant challenges for policy learning. The impulsive nature of cooperative jumping requires precise synchronization that is difficult to discover through random exploration alone. These difficulties are further exacerbated in a decentralized multi-agent setting, where each agent must infer its role from limited proprioceptive feedback while coping with a non-stationary environment shaped by the other’s actions. Moreover, discrepancies between simulation and reality introduce a substantial sim2real gap that jeopardizes real-world deployment. To navigate this complex learning landscape, we introduce a four-stage curriculum learning strategy that incrementally exposes the agents to increasing levels of task difficulty, thereby guiding exploration and enabling robust policy transfer. As illustrated in Fig.~\ref{pics:framework}b, our training procedure is structured sequentially into the following stages.

\subsection{Gravity Curriculum}
At the early stage of training, the agents lack prior knowledge of effective jumping postures. Even when the reward function encourages plausible motion patterns, the resulting jumps are typically too low to reach the target platform. Without successful task completions, training tends to converge to suboptimal policies, preventing the acquisition of high-performance behaviors. To alleviate this exploration bottleneck, we initially reduce gravitational acceleration, which allows higher jumps under lower gravity and increases the likelihood of reaching the platform—thereby providing informative reward signals for learning.

Specifically, gravitational acceleration is initialized at $7.0\,\mathrm{m/s^2}$ and incrementally increased to the nominal Earth gravity of $9.81\,\mathrm{m/s^2}$ at timesteps 15k, 20k, and 25k.

\subsection{Target Curriculum}
Directly exposing the agents to diverse jump targets—including varying heights and yaw orientations—from the beginning of training severely hinders exploration. In the absence of prior experience, the policy struggles to infer appropriate motor commands for directional jumps solely from sparse reward feedback. To mitigate this, we build upon the policy learned under the gravity curriculum and progressively expand the target space to improve generalization.

Training starts with a fixed target configuration ($h = 0.8\,\mathrm{m}$, yaw angle $\theta = 0^\circ$). Once the agents collectively achieve 25k successful jumps across all parallel environments, the maximum allowable yaw deviation $\theta_{\max}$ is increased by $15^\circ$ per curriculum phase until the full range $\theta \in [0^\circ, 90^\circ]$ is covered. Subsequently, the maximum jump height $h_{\max}$ is raised in $0.1\,\mathrm{m}$ increments from $0.8\,\mathrm{m}$ to $1.0\,\mathrm{m}$.

\subsection{Initialization Curriculum}
Initially, Robot J is placed above Robot L with a vertical gap along the $z$-axis between its feet and Robot L’s platform. This gap provides Robot J with a brief free-fall interval during which it can adjust its posture before initiating the jump, thereby facilitating the discovery of effective takeoff configurations. In contrast, initializing Robot J directly in contact with Robot L often causes premature takeoff and convergence to poor local optima. Ablation studies on this design choice are provided in Section～\ref{init}.

To systematically transition from this exploratory initialization to the final stacked configuration, the initial height $h_{\mathrm{init}}$ is linearly reduced from $1.0\,\mathrm{m}$ to $0.77\,\mathrm{m}$, while the body orientation evolves from an upright stance to a prone position resting flat on Robot L’s back. This transition is implemented via 15 linear interpolation steps between the two endpoint configurations, distributed across 15 curriculum phases to ensure smooth policy adaptation.

\subsection{Delay Curriculum}
In simulation, placing Robot J directly atop Robot L at episode start introduces instantaneous loading that excites transient oscillations in Robot L’s dynamics—effects that do not occur in the real world, where the stacked configuration is assembled statically. This sim2real discrepancy leads to inconsistent initial states at jump initiation, degrading transfer performance.

To align simulated initialization with real-world conditions, we enforce a settling delay $t_{\mathrm{delay}}$ at the beginning of each episode, during which both robots remain stationary (i.e., policy actions are clamped to zero) to allow mechanical vibrations to dissipate. The delay duration is progressively increased from $0\,\mathrm{s}$ to a final range of $[1.0, 1.6]\,\mathrm{s}$ over the course of training, thereby exposing the policy to realistic start-up conditions and enhancing robustness in physical deployment.

Additionally, to enable forward-flipping maneuvers, we incorporate a torque-assisted curriculum after the base jumping policy has been successfully acquired. Initially, a constant pitching torque of $120\,\mathrm{N\cdot m}$ is applied to Robot J during forward jumps to facilitate rotational motion. This auxiliary torque is then linearly decayed in steps of $4\,\mathrm{N\cdot m}$ until it reaches zero, ensuring that the policy gradually internalizes the necessary torque generation through its own actuation.

\section{Simulation Analysis}\label{Experiments}
\subsection{Simulation Setup}

\textbf{Environment and Agents.} We trained the policy across 4096 parallel environments in IsaacSim 4.5. The framework supports a quadruped pair: Robot L (Launcher) was instantiated as either the Unitree Aliengo or EngineAI Js01, while Robot J (Jumper) was instantiated as the Unitree Go1 or Go2. Episodes were terminated if the contact force on any robot base exceeded 1500 N.

\textbf{Curriculum Learning.} As detailed in Section~\ref{curriculum}, our base policy was trained on the Aliengo-Go1 platform with jump heights in the target curriculum ranging from $0.8\,\mathrm{m}$ to $1.0\,\mathrm{m}$.  The deployment policy for the Js01-Go2 pair was obtained by fine-tuning the pre-trained Aliengo-Go1 policy in two stages: first replacing Go1 with Go2 to adapt to the jumper's dynamics, then substituting Aliengo with Js01. In the final stage, the maximum jump height command was scaled to $h_{\text{max}} = 1.6\,\mathrm{m}$ to leverage Js01's superior payload capacity.

\subsection{Implementation Details}
Our MAPPO implementation leverages the framework provided by~\cite{yu2022surprising, serrano-munoz2023skrl}. Both the actor and critic networks employ identical four-layer Multilayer Perceptrons (MLPs) with hidden dimensions of $[512, 512, 256, 128]$ and Exponential Linear Unit (ELU) activations. The policy operates at a control frequency of $50\,\mathrm{Hz}$ to ensure real-time responsiveness. Training was conducted on a single NVIDIA RTX 4080 Super GPU. The complete five-stage curriculum required approximately 2.5 hours, with each stage averaging 0.5 hours.

To facilitate robust Sim-to-Real transfer, we applied extensive domain randomization as detailed in Table~\ref{rand_parameters}. These parameters are categorized into two groups: (i) intrinsic physical properties, governing the dynamics of individual agents; and (ii) inter-agent initialization states, which introduce stochasticity in the relative spatial alignment and temporal synchronization between robots.

\begin{table}[!htbp]
    \centering
    \caption{Domain Randomization Parameters}
    \vspace{0.5em}
    \label{rand_parameters}
    \footnotesize 
    \setlength{\tabcolsep}{3pt} 
    
    \begin{tabular}{llc}
        \toprule
        \textbf{Category} & \makecell{\textbf{Parameter} \\ \textbf{(Unit)}} & \makecell{\textbf{Range} \\ \textbf{[Min, Max]}} \\
        \midrule
        \multirow{10}{*}{\shortstack[l]{Robot Dynamics\\(Individual)}} 
         & Static friction & $[0.6, 1.0]$ \\
         & Dynamic friction & $[0.5, 0.9]$ \\
         & Push force disturbance (N) & $[-5.0, 5.0]$ \\
         & Push torque disturbance (N$\cdot$m) & $[-0.5, 0.5]$ \\
         & Actuator time lag (ms) & $[0, 10]$ \\
         & Center of mass offset (m) & $[-0.02, 0.02]$ \\
         & Motor stiffness gain & $[0.9, 1.1]$ \\
         & Motor damping gain & $[0.9, 1.1]$ \\
         & Added mass, Robot J (kg) & $[-2.0, 2.0]$ \\
         & Added mass, Robot L (kg) & $[-1.0, 1.0]$ \\
        \midrule
        \multirow{3}{*}{\shortstack[l]{Interaction\\(Inter-robot)}}
         & Communication delay (ms) & $[0, 5]$ \\
         & Rel. position offset (m) & $[-0.02, 0.02]$ \\
         & Rel. yaw offset (rad) & $[-0.08, 0.08]$ \\
        \bottomrule
    \end{tabular}
    \vspace{-1.0em}
\end{table}

\subsection{Simulation Results}

\begin{figure*}[!htbp]
   \centering
   \includegraphics[width=0.9\textwidth]{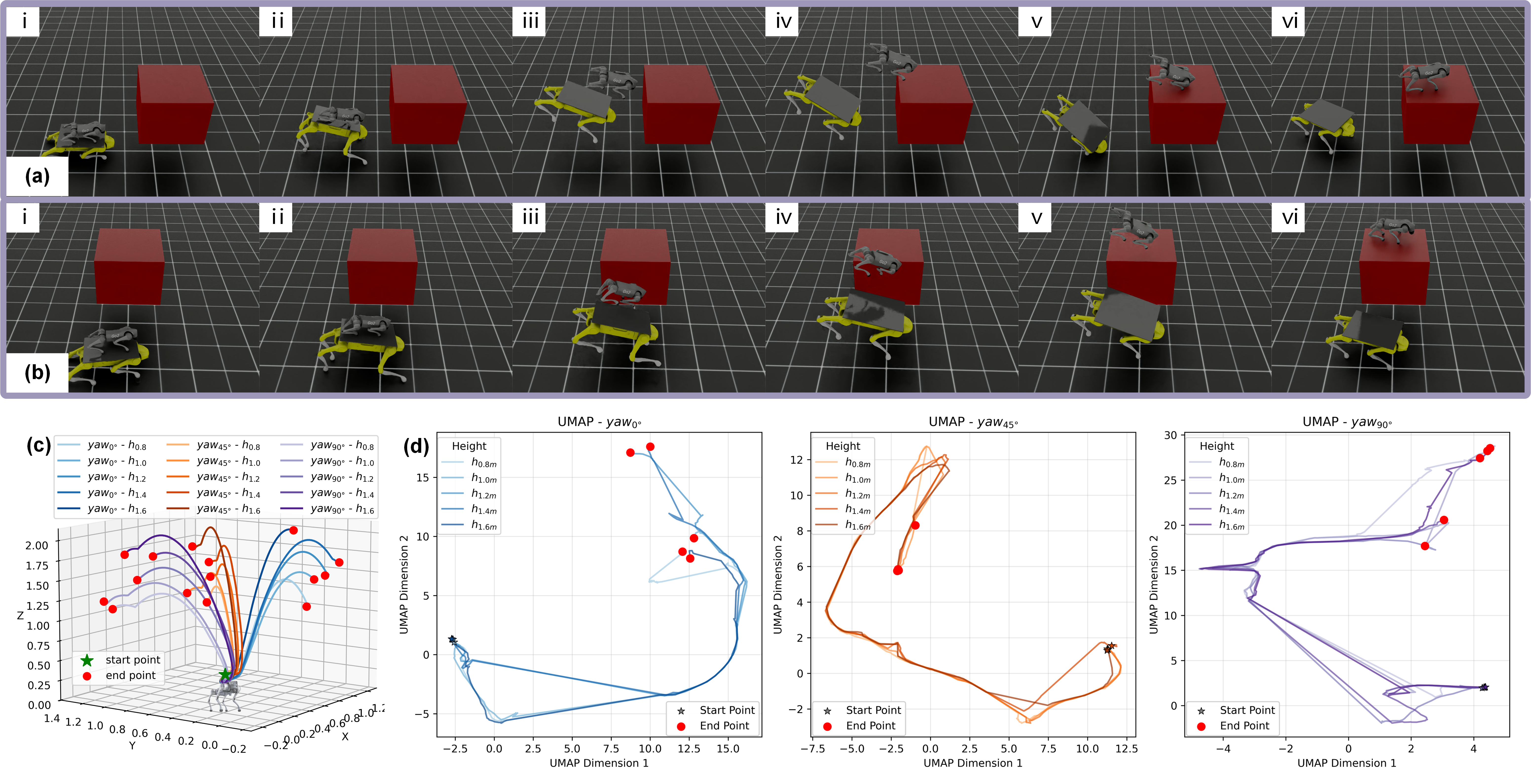}
   \caption{Simulation results of different jumping maneuvers, with Robot J jumping onto a $1.5\,\mathrm{m}$-high platform: (a) forward jump, and (b) side jump. These behaviors emerge from the progressive curriculum described in Section~\ref{curriculum}, (c)3D trajectories of successful jumps across varying target heights and yaw orientations. (d) UMAP embedding of joint command sequences, revealing distinct clusters corresponding to different maneuver types in the policy's latent action space.}
   \label{pics:sim_exp}
   \vspace{-1em}
\end{figure*}

The effectiveness of our curriculum learning strategy is demonstrated through a variety of cooperative jumping maneuvers executed by Robot J in simulation after full training. Specifically, Robot J can perform: (a) forward jumps (Fig.~\ref{pics:sim_exp}a), (b) side jumps (Fig.~\ref{pics:sim_exp}b), and (c) forward flips (Fig.~\ref{pics:cover}d).

\textbf{Trajectory Analysis.} We analyzed the motion patterns of Robot J by extracting root pose data across varying task commands, specifically yaw targets $\theta_{\text{yaw}} \in \{0^\circ, 45^\circ, 90^\circ\}$ and jump heights $h \in [0.8, 1.6]\,\mathrm{m}$. Fig.~\ref{pics:sim_exp}c illustrates the resulting trajectories, showing how the policy adapts to different targets.

\textbf{Motion Visualization.} To intuitively visualize the joint space of Robot J, we employed Uniform Manifold Approximation and Projection (UMAP)~\cite{huang2025learninga, mcinnes2020umap} to reduce the data into a 2D representation(Fig.~\ref{pics:sim_exp}d). The visualization reveals high structural similarity in jumping motions across tasks, with distinct variations occurring primarily during takeoff and landing. This indicates that the system executes a consistent jumping primitive, while locally modulating motion during ground contact to achieve different targets. Crucially, all task-specific adjustments are driven solely by changes to Robot J's commands, while Robot L maintains consistent behavior and supplies the majority of the vertical impulse. This shows that Robot J actively regulates the coupled dynamics to precisely control jump targets.

\subsection{Evaluation Metrics}

To quantitatively assess the performance and robustness of the proposed framework, we conducted evaluations across ten random seeds. Each seed involved 4096 parallel environments running for 20000 time steps. We employ the following quantitative metrics:

\begin{itemize} 
  \item \textbf{Sim-to-Real Transferability(\textbf{S2R})}: A binary indicator assessing whether the policy trained in simulation can be successfully deployed on physical hardware.
  \item \textbf{Success Rate ($E_{\text{Success}}$)}: The proportion of episodes where the Co-jump task is successfully completed. A trial is deemed successful only if Robot J satisfies a composite criterion: attaining sufficient vertical clearance, achieving high landing precision relative to the target, and maintaining an upright configuration.
  \item \textbf{Relative Height ($E_{h_{\text{diff}}}$)}: The instantaneous vertical distance between the two robots, defined as $h^{\text{J}} - h^{\text{L}}$.
  \item \textbf{Target Error ($\boldsymbol{E_{\epsilon_{\text{xy}}}}$)}: The Euclidean distance in the horizontal $xy$ plane between Robot J's base center and the designated target landing point.
  \item \textbf{Peak Jump Height ($E_{h^{\text{J}}_{\max}}$)}: The maximum vertical displacement of Robot J's base achieved during a single trajectory.
  \item \textbf{Power Consumption ($E_{\text{power}}$)}: $E_{\text{power}}$ is defined as the average absolute power across all joints: $E_{\text{power}} = \frac{1}{M} \sum_{i=1}^{M} |\boldsymbol{\tau}_i \cdot \dot{q}_i|$, where $\boldsymbol{\tau}_i$ and $\dot{q}_i$ are the torque and angular velocity of joint $i$, and $M$ is the total number of joints.
\end{itemize}

\subsection{Comparative Analysis}

\begin{figure*}[t]
   \centering
    \includegraphics[width=0.9\textwidth]{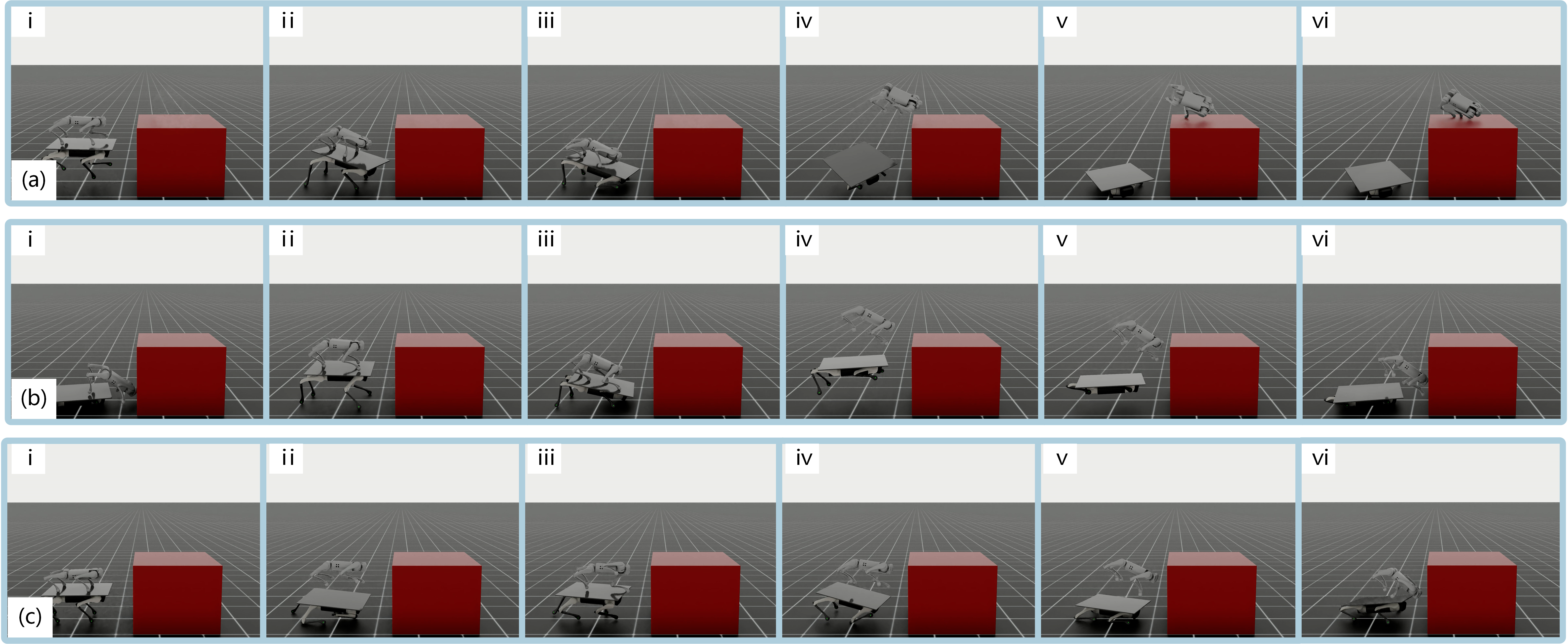}
   \caption{Ablation study results comparing different training configurations: (a) core curriculum training, (b) gravity curriculum ablation, (c) initialization curriculum ablation.}
   \label{pics:ablation}
   \vspace{-0.5em}
\end{figure*}

Given the absence of established multi-agent baselines for the Co-jump task, we benchmark our method against two state-of-the-art single-agent quadrupedal jumping frameworks: \textbf{Curriculum-Based Jumping}~\cite{atanassov2025curriculumbased} and \textbf{OmniNet}~\cite{han2025omninet}. We reimplemented both algorithms and deployed them independently on the Go2 and Js01 platforms without inter-agent communication. However, the Co-jump task inherently requires precise jump timing, exhibits strong dynamic coupling and contact instability during cooperative takeoff. Consequently, both baselines either fail catastrophically or produce unstable jumps. Ablation results are presented in Fig.~\ref{pics:ablation}.

\begin{table*}[t]
\centering
\caption{Quantitative comparison with baselines and ablation studies.}
\label{comparation_ablation}
\vspace{0.2em}
\setlength\tabcolsep{5pt}

\begin{tabular}{lcccccccc}
\toprule
\multirow{2}{*}{\textbf{Method}} & \textbf{Target} & \textbf{S2R} & $E_{\text{Success}}\uparrow$ & $E_{h_{\text{diff}}}\uparrow$ & $E_{\epsilon_{\text{xy}}}\downarrow$ & $E_{h^{\text{J}}_{\max}}\uparrow$ & \multicolumn{2}{c}{$E_{\text{power}}\downarrow$ (W)} \\
\cmidrule(lr){8-9}
 & $h_{\text{tgt}}$ (m) & & (\%) & (m) & (m) & (m) & Robot L & Robot J \\
\midrule

\multicolumn{9}{l}{\textit{\textbf{(a) Compared Methods} (Hardware: Js01-Go2)}} \\ 
\midrule
Atanassov et al.~\cite{atanassov2025curriculumbased} & 0.9 & \ding{55} & $0.01{\scriptstyle\pm 0.00}$ & $0.13{\scriptstyle\pm 0.19}$ & $1.11{\scriptstyle\pm 0.28}$ & $0.95{\scriptstyle\pm 0.06}$ & $95.3{\scriptstyle\pm 172}$ & $24.8{\scriptstyle\pm 21.7}$ \\
OmniNet~\cite{han2025omninet}                        & 0.9 & \ding{55} & $\symbf{98.1}{\scriptstyle\pm 1.71}$ & $0.25{\scriptstyle\pm 0.45}$ & $1.04{\scriptstyle\pm 0.64}$ & $1.54{\scriptstyle\pm 0.01}$ & $331.0{\scriptstyle\pm 208}$ & $33.6{\scriptstyle\pm 27.5}$ \\
\textbf{Ours}                                        & 0.9 & \ding{51} & $91.7{\scriptstyle\pm 5.29}$ & $\symbf{0.54}{\scriptstyle\pm 0.29}$ & $\symbf{0.28}{\scriptstyle\pm 0.35}$ & $\symbf{1.57}{\scriptstyle\pm 0.05}$ & $\symbf{54.5}{\scriptstyle\pm 106}$ & $\symbf{4.90}{\scriptstyle\pm 9.57}$ \\
\addlinespace[0.5em]
Atanassov et al.~\cite{atanassov2025curriculumbased} & 1.2 & \ding{55} & $0.00{\scriptstyle\pm 0.00}$ & $0.12{\scriptstyle\pm 0.19}$ & $1.11{\scriptstyle\pm 0.31}$ & $0.94{\scriptstyle\pm 0.05}$ & $95.2{\scriptstyle\pm 173}$ & $25.0{\scriptstyle\pm 21.8}$ \\
OmniNet~\cite{han2025omninet}                        & 1.2 & \ding{55} & $0.19{\scriptstyle\pm 0.01}$ & $0.11{\scriptstyle\pm 0.51}$ & $1.41{\scriptstyle\pm 0.46}$ & $1.54{\scriptstyle\pm 0.01}$ & $333.0{\scriptstyle\pm 210}$ & $37.4{\scriptstyle\pm 26.9}$ \\
\textbf{Ours}                                        & 1.2 & \ding{51} & $\symbf{92.8}{\scriptstyle\pm 5.36}$ & $\symbf{0.74}{\scriptstyle\pm 0.38}$ & $\symbf{0.25}{\scriptstyle\pm 0.28}$ & $\symbf{1.77}{\scriptstyle\pm 0.10}$ & $\symbf{56.7}{\scriptstyle\pm 114}$ & $\symbf{4.44}{\scriptstyle\pm 8.92}$ \\

\midrule
\multicolumn{9}{l}{\textit{\textbf{(b) Ablation Studies} (Hardware: Aliengo-Go1)}} \\
\midrule
Ours w/o gravity curriculum         & 0.8 & \ding{55} & $0.00{\scriptstyle\pm 0.00}$ & $0.25{\scriptstyle\pm 0.16}$ & $0.82{\scriptstyle\pm 0.05}$ & $1.06{\scriptstyle\pm 0.01}$ & $\symbf{4.45}{\scriptstyle\pm 14.0}$ & $17.3{\scriptstyle\pm 19.5}$ \\
Ours w/o initialization curriculum           & 0.8 & \ding{55} & $0.00{\scriptstyle\pm 0.00}$ & $0.34{\scriptstyle\pm 0.17}$ & $0.84{\scriptstyle\pm 0.22}$ & $0.70{\scriptstyle\pm 0.01}$ & $25.1{\scriptstyle\pm 18.9}$ & $16.2{\scriptstyle\pm 24.4}$ \\
\textbf{Ours (Base)}           & 0.8 & \ding{55} & $\symbf{97.7}{\scriptstyle\pm 2.08}$ & $\symbf{0.84}{\scriptstyle\pm 0.19}$ & $\symbf{0.17}{\scriptstyle\pm 0.02}$ & $\symbf{1.14}{\scriptstyle\pm 0.02}$ & $6.62{\scriptstyle\pm 15.6}$ & $\symbf{3.00}{\scriptstyle\pm 9.66}$ \\
\bottomrule
\end{tabular}
\vspace{-1.0em}
\end{table*}

 We analyze the performance across two difficulty levels to evaluate both the baseline capabilities and their scalability, as shown in Table~\ref{comparation_ablation}(a).

\textbf{Co-jump ($h_{\text{target}} = 0.9$\,m).}
At this moderate height, \textit{OmniNet} demonstrates a high success rate of $98.13\%$, suggesting that its policy can initially reach the target region. However, the average power consumption is exceptionally high ($E^\text{L}_{\text{power}} = 331.0$\,W; $E^\text{J}_{\text{power}} = 33.6$\,W) and exhibits severe post-landing instability. Robot J often falls off the platform due to poor landing states, which degrades $E_{h_{\text{diff}}}$ and further inflates energy costs.
\textit{Curriculum-Based Jumping}, hampered by severe temporal misalignment, fails to synchronize its takeoff phase with the launcher's motion. This timing disparity prevents effective momentum transfer, resulting in a negligible success rate ($0.01\%$) and large landing errors. 
In contrast, \textbf{Ours} achieves a comparable high success rate ($91.69\%$) but with fundamentally different motion characteristics. The energy consumption is an order of magnitude lower ($E^\text{L}_{\text{power}} = 54.5$\,W; $E^\text{J}_{\text{power}} = 4.90$\,W), and the landing precision is significantly tighter ($\epsilon_{\text{target}} = 0.28$\,m). This indicates that our method achieves success through precise coordination and efficient power transmission.

\textbf{Co-jump ($h_{\text{target}} = 1.2$\,m).}
As the task difficulty increases, the performance gap between methods widens. The success rate of \textit{OmniNet} drops sharply to $0.19\%$. This failure underscores that the high energy ($E^\text{L}_{\text{power}} = 333.0$\,W; $E^\text{J}_{\text{power}} = 37.4$\,W) input did not translate into correct cooperative gain.
\textit{Curriculum-Based Jumping} remains unable to complete the task.
In contrast, \textbf{Ours} maintains robust performance, with the success rate of $92.83\%$ and the peak height reaching $1.77$\,m. The system exhibits consistent stability and low energy usage, underscoring the necessity of centralized multi-agent training for successful cooperation in this task.

\subsection{Ablation Studies of Curriculum Design}

To evaluate the individual contributions of our curriculum strategies to the emergence of coordinated behaviors, we conduct ablation studies by selectively removing two core components during co-jump training at a target height of $h = 0.8\,\mathrm{m}$. As summarized in Table~\ref{comparation_ablation}(b), the quantitative results show that removing either component leads to a complete collapse in success rate.
\subsubsection{Ablation of the Gravity Curriculum}

When the gravity curriculum is removed and agents are trained directly under full Earth gravity, they completely fail to learn the task.
As shown in Table~\ref{comparation_ablation}(b), Robot J achieves a maximum jump height of $1.06$\,m, which is comparable to successful trials. However, the large target deviation ($E_{\epsilon_{\text{xy}}} = 0.82$\,m) indicates that the policy generates purely vertical lift but fails to steer the jump toward the target. Under full gravity, the feasible action space for successful cooperative jumping is extremely narrow.  Without curriculum-guided early successes, the policy cannot access the sparse target, get reaching reward and remains trapped in a suboptimal regime.

\subsubsection{Ablation of the Initialization Curriculum}\label{init}

Eliminating the initialization curriculum results in complete learning failure. In this setting, the policy converges to a degenerate, myopic strategy. Immediately upon episode reset, Robot J executes a premature, impulsive jump. This behavior greedily exploits the dense shaping rewards to maximize short-term returns, while completely disregarding the terminal objective of landing on the platform. The policy fails to explore the necessary temporal sequence required for the cooperative maneuver.

\section{Real-World Experiments}\label{Real_World_Experiments}
\subsection{Hardware Setup}

\textbf{Robotic Platforms.} We deployed two quadruped robots consisting of a Go2 as the jumper (Robot J, approx. 15\,kg) and a Js01 as the launcher (Robot L, approx. 90\,kg). To facilitate the launch, a $0.8\,\mathrm{m} \times 0.6\,\mathrm{m}$ wooden platform was securely mounted to the dorsal trunk of Robot L. Crucially, the system operated under a \textbf{blind setting}: both agents relied exclusively on onboard proprioception, with no external motion capture or inter-robot communication regarding state estimation.

\textbf{Control Architecture.} Real-time control was hosted on a centralized mobile workstation equipped with an AMD Ryzen 7 7435H CPU. Policy inference and action computation were executed on this host, which transmitted joint target position commands simultaneously to both robots via Ethernet to ensure precise temporal synchronization. The low-level PD gains were tuned to $K_p = 40$, $K_d = 1.2$ for Robot J and $K_p = 200$, $K_d = 5.0$ for the heavier Robot L.

\textbf{Experimental Environment.} The landing target was a modular workbench with a surface area of $1.0\,\mathrm{m} \times 1.0\,\mathrm{m}$. To evaluate the system's vertical scalability, the platform was configured to three distinct heights: $h \in \{0.9, 1.2, 1.5\}\,\mathrm{m}$. We also varied the target orientation ($0^\circ$ and $90^\circ$) relative to the launch vector. Across all trials, the horizontal standoff distance between the launcher and the target was fixed at approximately $1.2\,\mathrm{m}$.

\subsection{Experimental Results}

\begin{figure*}[!htbp]
   \centering
    \includegraphics[width=1.0\textwidth]{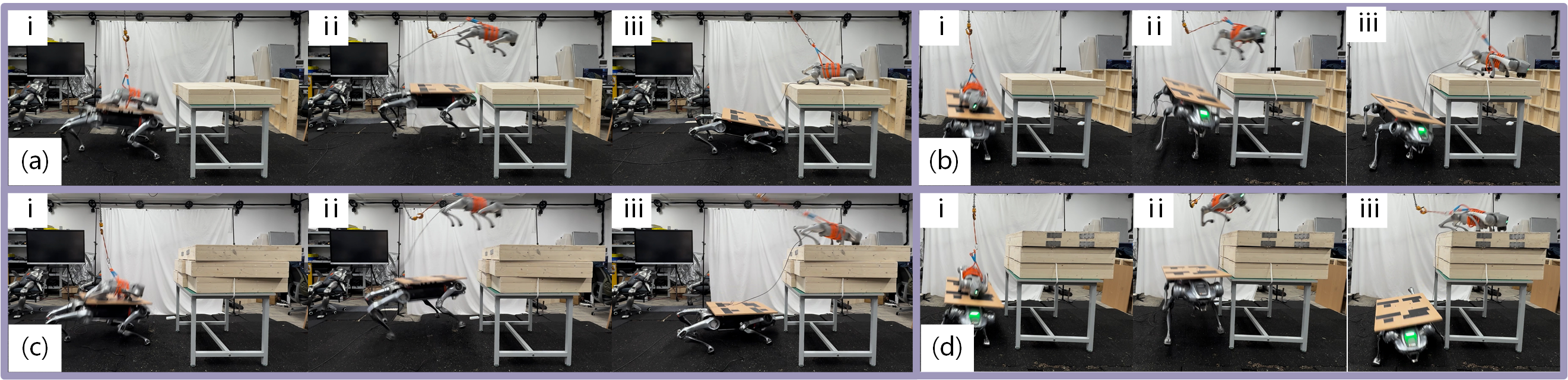}
   \caption{Snapshot sequences of real-world experiments showcasing different jumping maneuvers, including variable heights ($0.9$\,m and $1.2$\,m) and heading angles ($0^\circ$ and $90^\circ$).}
   \label{pics:real_maneuvers}
   \vspace{-0.5em}
\end{figure*}

\begin{figure*}[!htbp]
   \centering
   \includegraphics[width=1.0\textwidth]{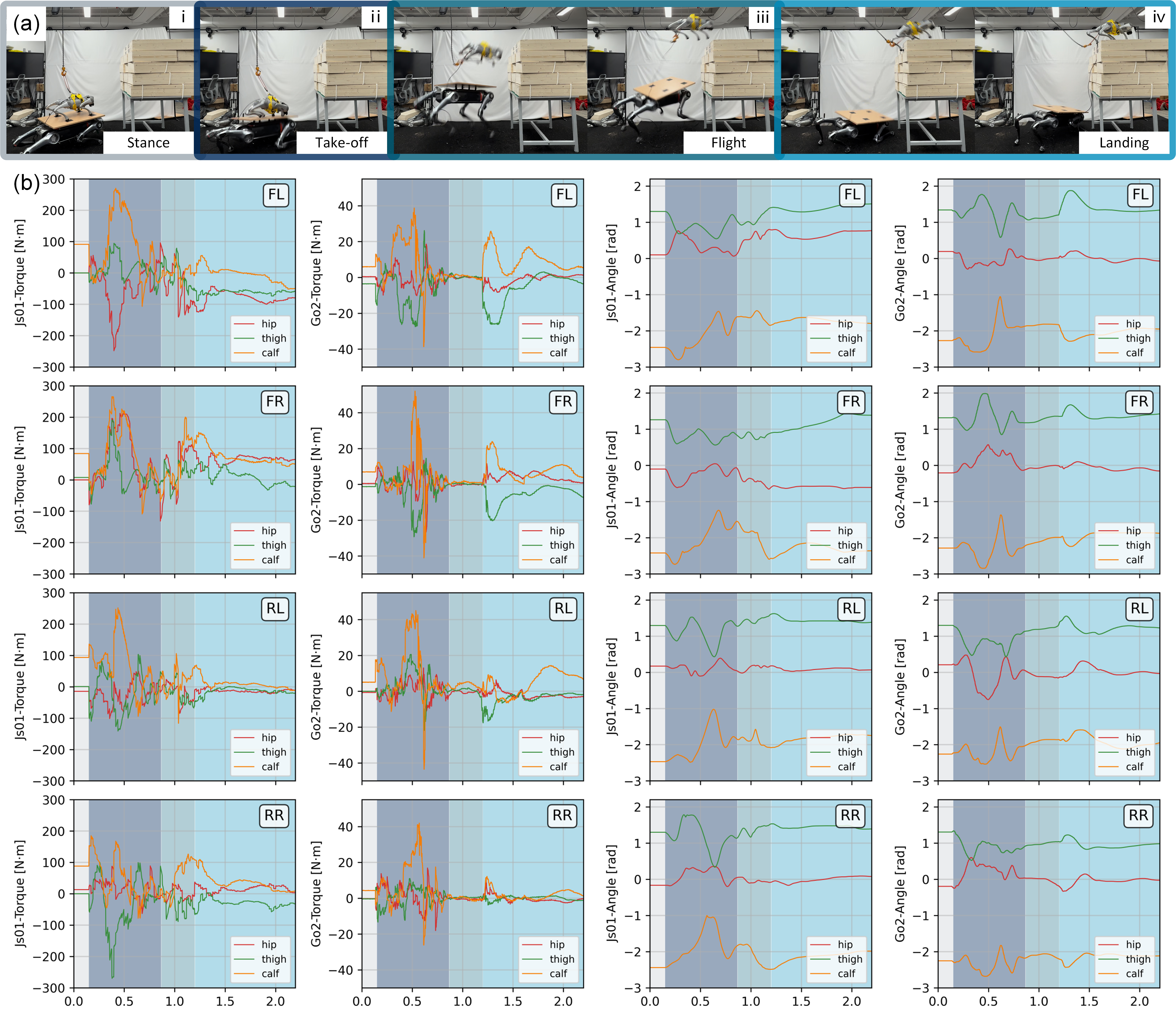}
   \caption{Detailed analysis of the $1.5$\,m cooperative jump. (a) Sequential snapshots of the experiment illustrating the four motion phases: Stance, Take-off, Flight, and Landing. (b) Real-world telemetry data showing the temporal evolution of joint torques and joint angles for both the Launcher (Js01) and Jumper (Go2).}
   \label{pics:exp-1.5m}
   \vspace{-0.5em}
\end{figure*}

\subsubsection{Overview of Deployment Capabilities}
To validate the system's versatility, we deployed the learned policy across a range of task configurations. The cooperative system successfully executed jumps to platform heights of $0.9$\,m, $1.2$\,m (Fig.\ref{pics:real_maneuvers}), and $1.5$\,m (Fig.\ref{pics:cover}c, Fig.~\ref{pics:exp-1.5m}). Additionally, the policy demonstrated robustness to directional variations, consistently performing both forward maneuvers ($\theta_{\text{yaw}} = 0^\circ$) and lateral maneuvers ($\theta_{\text{yaw}} = 90^\circ$).

\subsubsection{Kinematic and Dynamic Analysis}  
To provide deeper insights into the coordination mechanism, we analyze telemetry data from a $1.5$\,m forward Co-jump, representing the system's peak vertical performance. Fig.~\ref{pics:exp-1.5m} illustrates the time-lapse sequence along with the corresponding joint dynamics. The cooperative maneuver can be distinctly segmented into four dynamic phases:

\begin{enumerate}
    \item \textbf{Phase i: Stance}  
    At the start, the agents maintain a superimposed static equilibrium. Joint torques and angles remain nearly constant.

    \item \textbf{Phase ii: Take-off}  
    Upon triggering, the system transitions into the impulse generation phase, characterized by a rapid increase in joint torques for both robots. Robot L serves as the primary propulsion source, generating a significant vertical impulse to boost Robot J, while Robot J synchronizes its leg extension to maximize launch velocity.
    During the take-off phase, joint torque values for both robots approach their respective limits to provide sufficient propulsive force. Specifically, the peak torque in Robot L's calf joint reaches approximately 270\,$N \cdot m$, while Robot J's calf joint exceeds 40\,$N \cdot m$. In terms of timing, Robot L's maximum torque occurs around $t = 0.40$\,s, delivering the majority of the impulse. Subsequently, as Robot J receives sufficient kinetic energy, its torque peaks at $t = 0.56$\,s.

    \item \textbf{Phase iii: Flight}  
    Following lift-off, Robot J enters a ballistic flight phase. During this stage, Robot J exhibits negligible torque output, maintaining a fixed aerial posture to optimize its trajectory. Meanwhile, Robot L actively modulates its joints to dissipate recoil energy and prepares for its own controlled landing.

    \item \textbf{Phase iv: Landing}  
    In the final phase, Robot J successfully lands on the target platform. The torque data reveals sharp peaks corresponding to impact absorption, followed by compliant stabilization as Robot J restores its nominal standing posture. Concurrently, Robot L achieves a stable resting state, completing the cooperative maneuver.
\end{enumerate}

\subsection{Sim2Real Transfer Analysis}

Despite successful policy transfer from simulation to real hardware, several sim-to-real gaps limit performance. 

\textbf{Actuation constraints manifest in reduced torque output.} The simulated actuator model assumes idealized performance, whereas real-world execution exhibits lower torque during dynamic jumps. This discrepancy arises from unmodeled physical effects and operational constraints inherent to real hardware.

\textbf{Absolute jump height differs between domains.} After policy fine-tuning, our system achieves a maximum simulated jump height of $2.0\,\mathrm{m}$, whereas real-world experiments reach $1.5\,\mathrm{m}$. Consequently, to target real-world platforms at $\{0.9, 1.2, 1.5\}\,\mathrm{m}$, we issue higher simulated jump commands of $\{1.1, 1.4, 2.0\}\,\mathrm{m}$, respectively, to compensate for this systematic bias.

\textbf{Imperfect initial alignment introduces coordination errors.} In practice, the relative position and orientation of Js01 and Go2 cannot be perfectly replicated from simulation. Although domain randomization is applied during training to enhance robustness, excessive perturbations in relative pose degrade policy performance, indicating a bounded tolerance for initialization error. In real trials, small misalignments often lead to suboptimal takeoff postures for Robot J, resulting in unintended roll motion ($\varphi_{\text{roll}}$) during flight and unstable landings—even though the cooperative jumping motion itself remains reliably triggered.

\textbf{A critical sim2real mismatch stems from dynamic contact modeling.} During takeoff, Robot J transitions from a resting posture to a pre-jump stance, significantly shifting its center of pressure on Js01’s back platform. This repositioning alters the contact geometry in ways that are difficult to capture accurately in simulation, further contributing to roll instability. Such high-frequency, high-force interaction effects remain challenging to model faithfully across domains.

\section{Conclusions}\label{conclusions}

We have presented a novel learning-based framework that enables a quadruped pair to execute highly dynamic cooperative jumping in vertically constrained environments. By synergizing a multi-stage curriculum with explicit interaction-aware rewards, our method not only achieves high success rates in simulation but also demonstrates robust transfer to physical hardware without external perception. This work presents a pioneering approach to multi-agent legged locomotion in constrained, complex settings, demonstrating that coordination unlocks capabilities beyond the reach of individual agents.

\section{Limitations and Future Work}

While our approach yields compelling results, several limitations remain that suggest directions for future research.

\begin{itemize} \item \textbf{Sensitivity to Initial Conditions:} The current policy requires precise initial alignment between the robots. Deviations in the relative pose prior to takeoff can degrade performance or lead to failure. This sensitivity stems primarily from unmodeled dynamics in the sim2real gap, particularly regarding actuation bandwidth and contact friction, as well as the inherent challenge of coordinating high-impulse maneuvers under state uncertainty.

\item \textbf{Pre-assembled Initialization:} Our experiments assume the agents are initialized in a stacked configuration. This setup simplifies the problem by bypassing the navigational complexity of autonomous rendezvous and docking. A fully autonomous deployment would require a hierarchical planner capable of guiding the robots to approach, align, and mount one another before executing the jump.

\item \textbf{Scope of Maneuvers and Morphology:} This study focuses on a specific vertical jumping primitive using two quadrupedal agents. Future work could generalize this paradigm to more diverse morphological combinations, such as a humanoid robot leveraging a large quadruped as a mobile launch platform. Furthermore, the framework could be extended to encompass a broader repertoire of cooperative skills, including vaulting, climbing, and assisted recovery, to address a wider range of locomotion tasks.
\end{itemize}

\section{Acknowledgments}
This work was carried out during an internship at EngineAI. The authors gratefully acknowledge EngineAI for providing computational resources and technical guidance. 

\section*{APPENDIX}

\subsection{Experimental Details} \label{appendix:finetune}
The deployment policy for the Js01-Go2 pair was derived from the pre-trained Aliengo-Go1 baseline via a two-stage fine-tuning protocol. First, we substituted the Go1 model with the Go2 model to adapt the policy to the jumper's updated dynamics. Subsequently, we replaced the Aliengo with the Js01. Crucially, in this final stage, the maximum jump height command was scaled to $h_{\text{max}} = 2.0\,\mathrm{m}$, allowing the policy to fully exploit the superior payload capacity and actuation limits of the Js01 platform.


\subsection{MAPPO Hyperparameters}\label{appendix:mappo}
Details of our MAPPO implementation are summarized in Table~\ref{mappo_hyperparams}. The input observations are standardized by removing the mean and scaling them by the standard deviation:

\begin{table}[!htbp]
    \centering
    \caption{MAPPO Implementation Hyperparameters}
    \vspace{0.5em}
    \label{mappo_hyperparams}
    \footnotesize 
    \setlength{\tabcolsep}{3pt} 
    \begin{tabular}{lc}
        \toprule
        \textbf{Hyperparameter} & \textbf{Value} \\
        \midrule
        Log-std range & $[-20.0, 2.0]$ \\
        Rollout steps & 16 \\
        Learning epochs & 5 \\
        Mini-batches & 4 \\
        Discount factor ($\gamma$) & 0.99 \\
        GAE balancing factor ($\lambda$) & 0.95 \\
        Learning rate & $5.0 \times 10^{-4}$ \\
        KL-Adaptive threshold & 0.016 \\
        Gradient norm clip & 1.0 \\
        Ratio clip & 0.2 \\
        Value clip & 0.2 \\
        \bottomrule
    \end{tabular}
    \vspace{-1.0em}
\end{table}

\bibliographystyle{plainnat}
\bibliography{root}

\end{document}